\documentclass[conference]{./IEEEtran}

\usepackage{epsfig}
\usepackage{times}
\usepackage{amsmath}
\usepackage{amssymb}
\usepackage{graphicx}
\usepackage{cite}
\usepackage{subcaption}
\usepackage[]{algorithm2e}
\usepackage{framed}

\usepackage{multirow} % Required for multirows
\usepackage{booktabs} % For prettier tables

\begin{document}

\title{New Wing Stroke and Wing Pitch Approaches for Milligram-scale Aerial Devices} 

\author{\IEEEauthorblockN{Palak Bhushan and Claire Tomlin} 
\IEEEauthorblockA{Dept. of EECS, UC Berkeley. 
}
}

\maketitle

\begin{abstract}
Here we report the construction of the simplest transmission mechanism ever designed capable of converting linear motions of any actuator to $\pm$60$^\circ$ rotary wing stroke motion. It is planar, compliant, can be fabricated in a single step and requires no assembly. Further, its design is universal in nature, that is, it can be used with any linear actuator capable of delivering sufficient power, irrespective of the magnitude of actuator displacements. We also report a novel passive wing pitch mechanism whose motion has little dependence on the aerodynamic loading on the wing. This exponentially simplifies the job of the designer by decoupling the as of yet highly coupled wing morphology, wing kinematics and flexure stiffness parameters. Like the contemporary flexure-based methods it is an add-on to a given wing stroke mechanism. Moreover, the intended wing pitch amplitude could easily be changed post-fabrication by tuning the resonance mass in the mechanism. 
\end{abstract}

\IEEEpeerreviewmaketitle

%%%%%%%%%%%%%%%%%%%%%%%%%%%%%%%%%%%%%%%%%%%%%%%%%%%%
\section{Introduction}
Flapping wing microrobots \cite{wood_liftoff, lin_EM1, lin_EM2, lin_electrostatic, zhang16, baybug18} utilize insect wing kinematics to generate lift, which have periodic 80$^\circ$-150$^\circ$ peak-to-peak wing stroke and approximately 90$^\circ$ peak-to-peak wing pitch amplitudes \cite{passive_rot}. 
Most contemporary transmission designs use the mechanical advantage principle to amplify small actuator displacements \cite{actuator_selection} into large wing stroke rotations. This operating principle in conjunction with the small linear displacements involved demand a high resolution requirement on the fabrication methods used \cite{baybug19}. Moreover, due to the macro-scale mechanism design principle of rigid-links plus revolute-joints used, these are typically a multi-material design followed by 3D assembly. Here, inspired from \cite{baybug18}, we use a resonant transmission instead to achieve large wing strokes which relaxes the resolution demand on the fabrication procedure used. We also design this resonant system to be planar and single-material for ease in assembly. 

There is a 90$^\circ$ phase difference between the stroke and pitch in order to always maintain a positive wing angle-of-attack. For simplicity in fabrication all works in the milligram-size-scale only actuate stroke and achieve wing pitch passively by utilizing the 90$^\circ$ phase-lagged (with respect to stroke) aerodynamic loading on the wing together with a flexure joint of appropriate rotational stiffness near the wing leading edge and operate the system quasi-statically \cite{passive_rot}. 
Not only is the flexure manufacturing elaborate involving aligned sandwiching of multiple laser cut layers \cite{endurance}, tedious design iterations occur due to the non-linear interdependency between flexure stiffness, wing pitch, wing shape and size caused by complex fluid dynamics. Different stiffness hinges needed to be tried for any changes made to the wing shape and to achieve desired pitch. Here we use centripetal force as opposed to wing damping to produce fairly decoupled 90$^\circ$ phase-lagged wing pitch, and use better fatigue strength metals like steel as opposed to polymers \cite{endurance}. 
This is also easier to manufacture and manually assemble because of a mostly single-material planar design. 

\begin{figure} 
\centering 
\epsfig{file=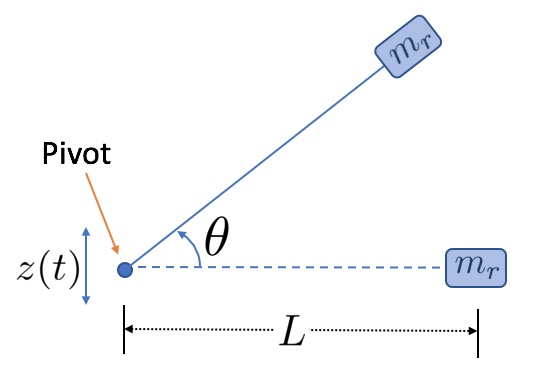,width=1.8in}
\caption{{Linearly driven torsional pendulum with linear excitation in the marked z-direction. }} 
\vspace{-1.5em}
\label{fig:torsional_pendulum}
\end{figure} 

%%%%%%%%%%%%%%%%%%%%%%%%%%%%%%%%%%%%%%%%%%%%%%%%%%%%
\section{Design} 
\subsection{Stroke mechanism physics}
A simple pendulum is able to convert linear motion of the support to rotary motion of the mass. We use this idea and replace the restoring force due to gravity by a torsional pivot, thus creating a linearly driven torsional pendulum like the one shown in Fig. \ref{fig:torsional_pendulum}. 

\begin{figure}
\hspace{-1.5em}
\epsfig{file=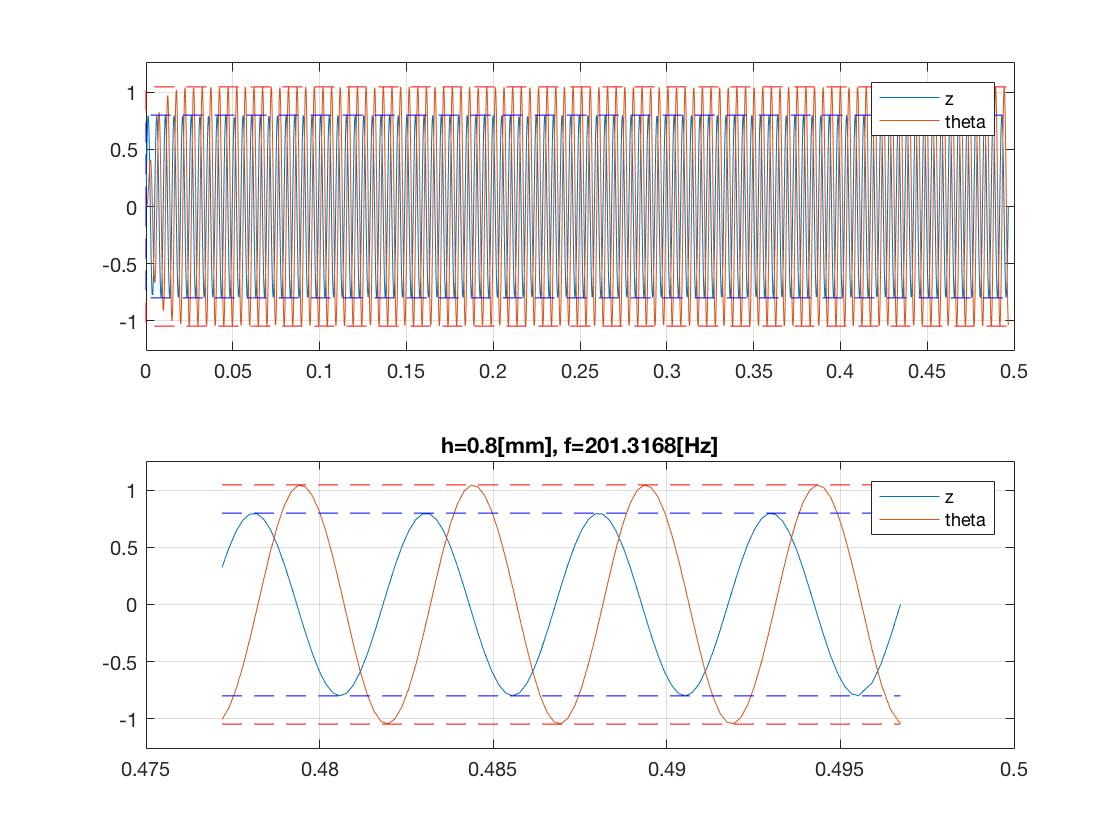,width=3.5in}
\vspace{-2.0em}
\caption{{MATLAB simulation of ODE governing linearly driven torsional pendulum. For the height $z(t)$ of the pivot oscillating with amplitude of 0.8mm, an angular oscillation amplitude of 60$^\circ$ is achieved by the torsional spring.}}
\vspace{-1.5em}
\label{fig:simple_driven_torsional_pendulum}
\end{figure}

This linearly driven torsional pendulum is governed by
\begin{equation}
I_x \ddot \theta = -k_t \theta - bL_w^2\dot \theta -bL_w\dot z \cos(\theta) -m_rL\cos(\theta)\ddot z
\end{equation}
where the pivot is driven periodically as $z(t) = z_{max} \sin(\omega t)$, and thus $\dot z(t)$, $\ddot z(t)$ are known functions of time. 
$m_r$ is the mass added at radial distance $L$ for resonance, $I_x=m_rL^2$ is the inertia of the rotating mass about rotation axis, $k_t$ is torsional spring constant, $L_w$ is the radial distance of the wing's effective center-of-pressure (c-p) from the pivot, and $b$ determines the effective angular damping acting at $L_w$. $\omega$ is chosen $=\sqrt\frac{k_t}{I_x}$ for resonance. For a 100mg aerial device, assuming 3x average lift force $\approx$ 1.5mN to be the peak force seen by each of the two wings, the damping factor $b$ is chosen as 
\begin{equation}
b = \frac{1.5 \cdot 10^{-3}}{L_w \omega \frac{\pi}{3} + z_{max}\omega}
\end{equation}
to approximately mimic wing damping. Let us assume some representative values for the actuator displacement $z_{max}$ and its frequency $f$, the resonant mass $m_r$ and wing's $L_w$. For $z_{max}$ = 0.8mm, $f$ $\approx$ 200Hz, $m_r$ = 2mg, and $L_w \approx$ 4.4mm, MATLAB differential equation simulation yields the following working values for the rest of the parameters 
\begin{equation}
L=2.5\text{mm}, ~k_t=20\mu\text{Nm}
\end{equation}
in order to achieve stroke angle of $\pm$60$^\circ$ (see Fig. \ref{fig:simple_driven_torsional_pendulum}). 
%These same parameters work for $f$ $\approx$ 70Hz and $m_r$ = 18mg as well because of scaling. 
Appropriately scaled values of $f$ and $m_r$ work as well. 
Note that the above is a simplified model capturing the basic idea of the transmission, and the actual transmission will have a more distributed resonant mass in contrast to the point mass $m_r$. 

The inertial and elastic forces and torques of the order of 10mN and 20$\mu$Nm, respectively, dominate over the damping forces and torques of the order of 1mN and 5$\mu$Nm due to drag and lift in the proposed design. Thus, while designing for the stiffness of the springs, it is sufficient to just consider the inertial forces and torques which are given by
$F_{inertial} = m_{r} \omega^2 L \approx 8\text{mN},$
and, $T_{inertial} = F_{inertial} \cdot L \approx$ 20$\mu$Nm. Thus, an inertial torque of 20$\mu$Nm acting on a $k_t$ = 20$\mu$Nm torsional spring will cause an angular displacement of 1rad $\approx \frac{\pi}{3}$rad = 60$^\circ$, just as desired. 

\subsection{Compliant pivot}
\begin{figure}
\epsfig{file=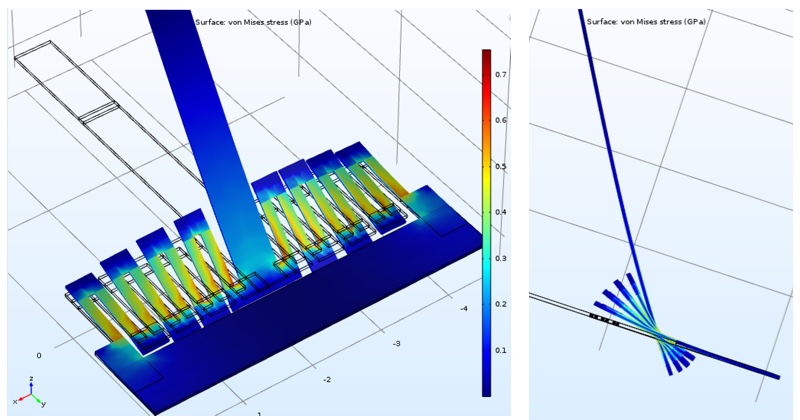,width=3.3in}
\caption{{Compliant pivot capable of large rotations with small stresses. x-axis is the pivot's rotation axis. This design achieves a rotation of 60$^\circ$ with an applied torque of $T_x$ = -20$\mu$Nm. All the stresses during operation are well below 0.8GPa which cold rolled steel can withstand. 
}}
\vspace{-1.0em}
\label{fig:pivot}
\end{figure}

The compliant planar spring (see Fig. \ref{fig:pivot}) is designed according to the method outlined in \cite{baybug18, baybug19} and laser cut from 301 stainless steel (see Table 1). 

\begin{table}[h!]
\normalsize
  \centering 
    \caption{Steel spring specifications.}
    \label{tab:spring-specs}
    \begin{tabular}{|l|r|}
    \hline
      \# parallel beams & 16 \\
      \hline
      Length of each beam & 1mm \\
      \hline 
      Beam width & 0.1mm \\
      \hline 
      Beam thickness & 38um \\ 
      \hline
    \end{tabular}
    \vspace{-1.4em}
\end{table}

\subsection{Pitch mechanism physics}

\begin{figure}
\hspace{-0em}
\epsfig{file=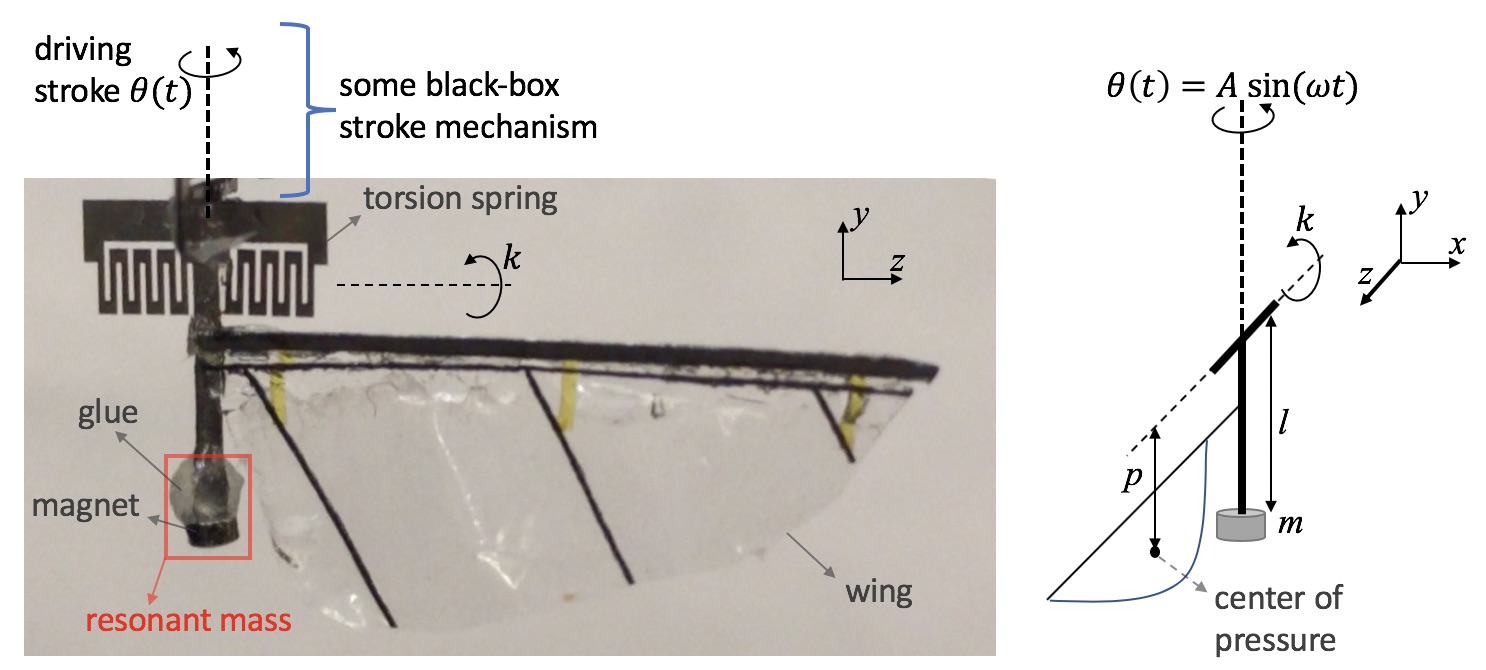,width=3.4in}
\vspace{-0em}
\caption{{(Left) Front view of the assembled pitch mechanism with 1mg steel spring and 0.4mg 15mm wing. (Right) Perspective view of the abstract mechanism in neutral wing stroke and neutral wing pitch position. }}
\vspace{-1em}
\label{fig:p1}
\end{figure}

\begin{figure}
\hspace{-0em}
\epsfig{file=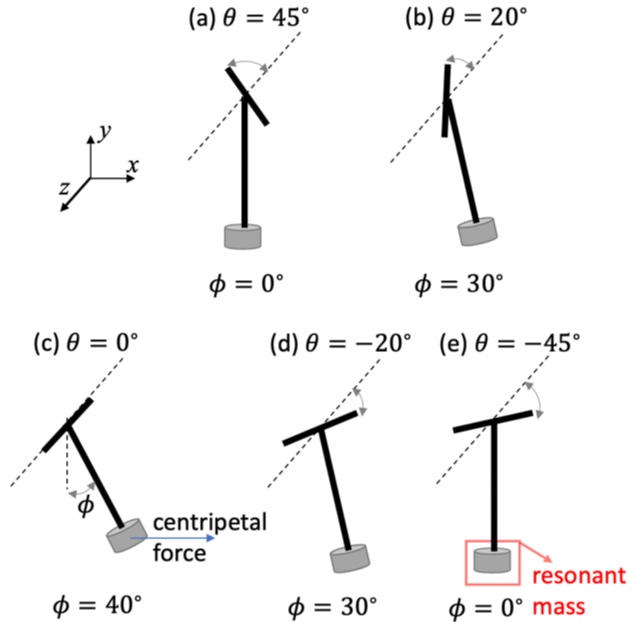,width=2.6in}
\vspace{-0em}
\caption{{Perspective view of the abstract mechanism while in motion from the positive extreme to the negative extreme stroke angle. These correspond to the last 5 snapshots in Fig.\ref{fig:p4}. }}
\vspace{-1em}
\label{fig:p3}
\end{figure}

The proposed wing pitching mechanism is driven with a periodic stroke angle externally (Fig. \ref{fig:p1}). We utilize the fact that centripetal forces are maximum at zero stroke (maximum stroke speed) and zero at extreme stroke angles (zero stroke speed) (Fig. \ref{fig:p3}). Wing is made using carbon-fiber veins and polyester membrane in a process similar to
\cite{wood_liftoff07, baybug18, baybug19}. A $m=$ 4mg magnet is glued at a distance $l$ from the torsion spring’s rotation axis. System's motion is described by
\begin{equation}
\begin{split}
ml^2\ddot \phi = & -k\phi + b L_w A\omega \cos(\omega t) p \\ 
+ & ml\sin(\phi) (A\omega\cos(\omega t))^2 l\cos(\phi) 
 \end{split}
\end{equation}
where distance $p$ of the wing’s c-p from z-axis is estimated at 2.5mm, distance $L_w$ from y-axis is estimated at 4mm, and $b$ is the effective damping coefficient of the wing such that $b L_w A\omega$ = 1mN to be approximately consistent with the peak aerodynamic force seen by the wing. During motion the spring is bent due to the torques caused by the radial centripetal forces in the xz-plane acting on the mass, and due to the aerodynamic force acting at wing’s c-p. Using ODE simulations in order to achieve $\phi=\pm45^\circ$ we select $l$ = 5mm and $k$ = 20$\mu$Nm, and use the same spring as before. 

%%%%%%%%%%%%%%%%%%%%%%%%%%%%%%%%%%%%%%%%%%%%%%%%%%%%
\section{Experiments} 
\begin{figure}
\epsfig{file=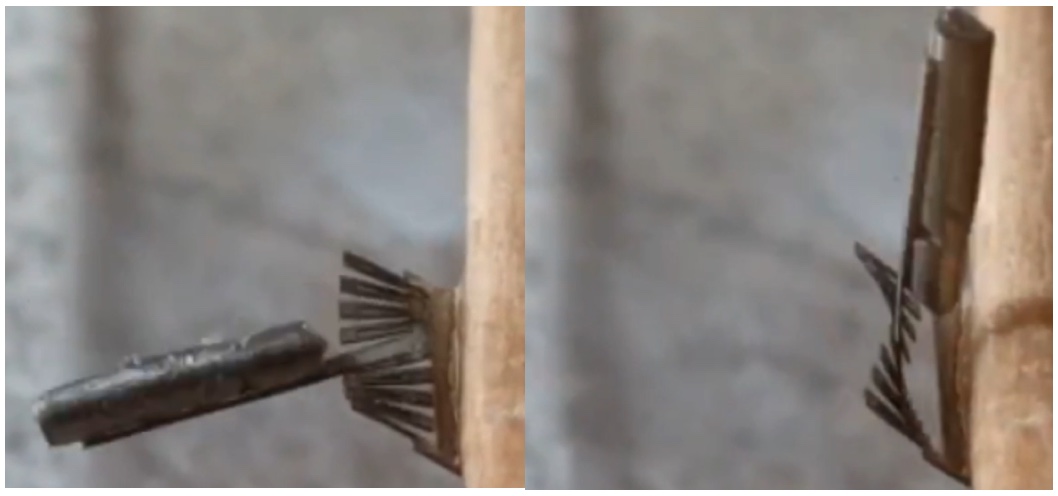,width=3.0in}
\caption{{Transmission excited via an external vibrating bench (seen in brown). }}
\vspace{-1.5em}
\label{fig:transmission_120deg}
\end{figure}

To demonstrate the stroke mechanism we excite the transmission at 70Hz without a damper using an external linearly vibrating bench and observe the rotation amplitude possible. $m_r$ is tuned to $\approx$ 18mg to observe resonance. Fig. \ref{fig:transmission_120deg} shows amplitudes exceeding $\pm$60$^\circ$, and a fixed axis of rotation, thus verifying our transmission design. 

\begin{figure}
\hspace{-0em}
\epsfig{file=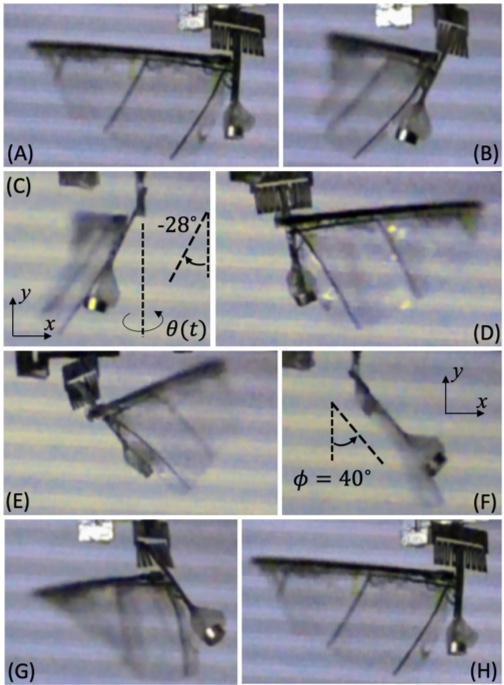,width=3.0in}
\vspace{-0em}
\caption{{Side view (xy-plane) of the mechanism while in motion for a complete wing stroke cycle, starting and ending at the negative extreme stroke angle. The last 5 snapshots (D-H) correspond conceptually to the 5 positions described in Fig. \ref{fig:p3}. Maximum wing pitch occurs at mid-stroke (see (F)), and zero pitch occurs at the 2 extreme strokes (see (A=H) \& (D)). }}
\vspace{-1em}
\label{fig:p4}
\end{figure}

Next, the pitch mechanism is driven at $\pm45^\circ$ stroke and $f$ = 70Hz. The resonant mass is tuned using glue and attaching smaller magnets till a decent wing pitch amplitude is observed (Fig. \ref{fig:p4}). From equation (4), the peak aerodynamic torque (2nd term) on the spring is estimated at $bL_wA\omega p = 2.5$uNm and the peak inertial torque due to the mass (3rd term) is estimated at 8uNm portraying the dominance of inertial loading over aerodynamic loading. Other potential benefits of this design is it could support heavier wings and could be made to function in other mediums like water. 

%%%%%%%%%%%%%%%%%%%%%%%%%%%%%%%%%%%%%%%%%%%%%%%%%%%%
\section{Conclusion} 
We presented a resonant transmission mechanism that can amplify the periodic displacements of any actuator into large rotations. To achieve the complete wing kinematics we then designed a passive wing pitch mechanism than can be driven by the first stroke mechanism, and whose motion is additionally decoupled from the wing loading. 

%%%%%%%%%%%%%%%%%%%%%%%%%%%%%%%%%%%%%%%%%%%%%%%%%%%%
\section*{Acknowledgements} 

The authors are grateful to get support from Commission on Higher Education (award \#IIID-2016-005)
and DOD ONR Office of Naval Research (award \#N00014-16-1-2206). 
We would also like to thank Prof. Ronald Fearing for his help and insightful discussions.

\end{document}